# Unleashing the Power of Transfer Learning Model for Sophisticated Insect Detection: Revolutionizing Insect Classification

[1] Md. Mahmudul Hasan, [1] SM Shaqib, [1] Ms. Sharmin Akter, [1] Rabiul Alam, [1] Afraz Ul Haque, [1] Shahrun akter khushbu

[1] Department of Computer Science and Engineering, Daffodil International University, Dhaka 1341, Bangladesh

**Abstract**

**Background** The purpose of the Insect Detection System for Crop and Plant Health is to keep an eye out for and identify insect infestations in farming areas. By utilizing cutting-edge technology like computer vision and machine learning, the system seeks to identify hazardous insects early and accurately. This would enable prompt response to save crops and maintain optimal plant health.

**Methods** The Method of this study includes Data Acquisition, Preprocessing, Data splitting, Model Implementation and Model evaluation. Different models like MobileNetV2, ResNet152V2, Xecption, Custom CNN was used in this study.

**Results** In order to categorize insect photos, a Convolutional Neural Network (CNN) based on the ResNet152V2 architecture is constructed and evaluated in this work. Achieving 99% training accuracy and 97% testing accuracy, ResNet152V2 demonstrates superior performance among four implemented models. The results highlight its potential for real-world applications in insect classification and entomology studies, emphasizing efficiency and accuracy.

**Conclusion** To ensure food security and sustain agricultural output globally, finding insects is crucial. Cutting-edge technology, such as ResNet152V2 models, greatly influence automating and improving the accuracy of insect identification. Efficient insect detection not only minimizes crop losses but also enhances agricultural productivity, contributing to sustainable food production. This underscores the pivotal role of technology in addressing challenges related to global food security.

Correspondence: shaqib15-4614@diu.edu.bd

# Background

The Food and Agriculture Organization (FAO) estimates that agricultural pests cause a 20–40% yearly reduction in crop productivity globally [1]. Farmers are primarily concerned with minimizing crop loss due to pests and diseases, as these issues arise regardless of the farming method employed [2].A significant component of forest ecosystems are forest insects [3].In today's medical industry, herb plants are vital and beneficial to humans [4]. Crop growth is substantially affected by insect calamities, exerting a notable influence on crop production. [5]. Numerous plant pests exist that are widely dispersed, quickly reproduce, and produce huge quantities of damage to crops [6]. By preventing these damages, a significant amount of the harvest may be saved, and agricultural productivity could increase [7]. Ensuring the safety of food requires the detection and identification of insects [8]. Recognition and identification of insect pests are essential to guarantee a good level of living, a healthy agricultural economy, and food security [9]. Using strips of yellow sticky paper is the most popular method for keeping an eye on insect problems. Without the use of any machinery or gadget, the insects imprisoned on these yellow sticky sheets are often counted by hand [10]. The identification and counting of the insect pests caught in the traps is a labor-intensive task. [11]. Identifying and classifying insects takes a lot of time and specialized knowledge for integrated pest Management in orchards [12]. For a long time, crop technicians and seasoned farmers were the primary sources of information for pest identification based on morphology [13]. Identifying insect pests from photos is a significant and difficult research problem [14]. Due to the modest resolution of insects in a picture and additional annoyances like occlusion, noise, and lack of features, handling the detection of microscopic items in large datasets can be difficult [15]. Digital technology and artificial intelligence approaches are widely used to automate many operations in several application domains, including smart transportation, industry, and health. It is also possible to use these methods in agriculture [16]. By encouraging the public to use smartphones to report bug sightings, it is possible to expand the scope of conventional trap monitoring [17].

The advantages of Machine Learning algorithms are their excellent image identification accuracy and well-defined structure [18]. Utilizing machine learning models can assist in making cost-effective decisions for crop protection, as opposed to the labor-intensive, time-consuming, and expensive nature of traditional monitoring methods. [19]. To successfully stop the invasion of unidentified species, there is still tremendous space for improvement in the precision of insect identification [20]. Wenyong Li et al. [21] propose a system to classify insects or pests using deep learning algorithm named Smart Pest Monitoring (SPM). The motive for doing this is the impact of pests on farmers' cropland and food quality. CNN algorithm achieved the best accuracy. Zeba Anwar et al. [22] identified insect using deep learning algorithms. In this case, ensemble-based models that make advantage of transfer learning are used for categorization. Here, a pre-trained model is employed. using Inception v3, Xception, VGG19, VGG16, and ResNet50 as voting classifier ensemble techniques. The focus of this paper revolves around the IP 102 dataset, encompassing 75,222 images distributed across 102 classesWhen the ensemble model was applied to the dataset, it showed an accuracy of 82.5%. Nour Eldeen M. Khalifa et al. [23] employs a transfer learning model to depict insect recognition. For the task at hand, the study selects deep transfer learning models, namely AlexNet, GoogleNet, and SqueezeNet. By using data augmentation techniques, overfitting-related problems are effectively addressed and models are improved. The AlexNet model showed the highest accuracy of all the models that were discussed, reaching approximately 89.33%. Mohammad Kamran Khan et al. [24] uesd InceptionV3 and VGG19, two deep transfer learning models, are used to identify regions containing different features. The highest accuracy of 81.7% and 80% was obtained using inceptionV3 and VGG19 models respectively. Yiwen Liu et al. [25] used multilayer network model to detect crop pests. This work is done in several steps such as image collection, providing a dataset of recognized models etc. VGG16 and Inception-ResNet-v2 have been applied for this task. Also, the model for insect image recognition and analysis combines two CNN series to enhance overall accuracy. The research was conducted with the assistance of the IDAP dataset. The suggested model for pest detection exhibited an accuracy rate of 97.71%. A paper by Wei Li et al. [26] proposes the application of YOLOv5, Mask-RCNN, and Faster-RCNN for crop pest identificationBased on the IP102 dataset and the Baidu AI Pests & Insects dataset, two cocoa datasets are created. Between the two datasets, the IP102 dataset yielded superior accuracy, with Fast-RCNN and Mask-RCNN achieving the highest accuracy at approximately 99%, while YOLOv5 exhibited a slightly lower accuracy of 97%. Ana Cláudia Teixeira et al. [27] emphasized two challenges and corresponding recommendations, specifically addressing issues related to dataset characteristics and methodology. João Gonçalves et al. [28] employed a dataset that included 168 photos taken using mobile devices from yellow sticky and delta traps, covering 8966 important insects. The SSD ResNet50 model performed best when it was deployed on mobile devices, with F1 scores ranging from 58% to 84% and per-class accuracies between 82% and 99%.

To improve machine learning models' prediction accuracy, Ming-Fong Tsai et al. [29] used time-series feature extraction is combined with transfer learning technology. And according to this research, there is a minimum 3% and a maximum 15% boost in prediction accuracy. Chen Li et al. [30] suggest a study that detects insects using a transfer learning convolutional neural network. The agricultural insect photo dataset from IP102 served as the source of data for this project. The ResNeXt-50 (32 × 4d) model is used in this study for classification; the combination of cutmix, fine-tuning, and transfer learning produced the highest classification accuracy of 86.95%. To detect insects, Ahmad Iftikhar et al. [31] use eight models, including several YOLO object detection architectural branches including YOLOv5, YOLOv3, YOLO-Lite, and YOLOR. With the best accuracy, the YOLOv5x model can identify 23 different insect species in 40.5 milliseconds. Applying the model to the IP-23 dataset yielded an average precision of 98.3%, mean precision of 79.8%, precision of 94.5%, recall of 97.8%, and F1-score of 97%. Xu Cao et al. [32] used VGG16, VGG19, InceptionV3 and InceptionV4 models on image processing and ImageNet dataset to detect insects. Among all the models VGG19 using Convolutional Neural Network obtained the highest accuracy i.e. 97.39%. VGG19 uses convolutional neural network method with highest accuracy, simple and convenient, less time consumption. Tuan T. Nguyen et al. [33] uses an efficient model of deep learning to classify insects. The model is based on EfficientNet to classify insects. All models trained over 10 epochs yielded the highest precision of 95% for all criteria, precision, recall, precision and F1-measure. EfficientNet-b3 gave the highest performance in this task. Catherine M. Little et al. [34] discussed aboutboth bad aspects such as destruction of agricultural and forest crops, disease of agricultural land as well as small organisms, as well as good and beneficial aspects such as biological control, contribution to plant and disease vectors, pollination. Different species of insects like small borer, good bad etc. are discussed. Kim BjergeID et al. [35] used a dataset contains 29,960 annotated insect images. This paper uses the YOLO algorithm to detect and classify insects. The YOLOv5 model had the highest results among all the model clusters and identified nine insect species that play a role in pollination. The model's average recall was 93.8% and accuracy was 92.7% when it came to identifying and categorizing insects and pests. In their research article, Eray Önler et al. [36] primarily use artificial intelligence algorithms and data science to identify thistle caterpillars (Vanessa cardui). This work use object detection in video to identify insects using the YOLOv5 object detection architecture of deep learning algorithm. 2416 images are used for this model. Here the scratch method and transfer learning model for object detection are trained and the results are presented. Michael Gomez Selvaraj et al. [37] used DCNN and Transfer Learning Algorithm. It features over 18,000 pre-screened banana images and over 30,952 annotations from farmers in Africa, Latin America and South India. The models obtained between 70 and 99% accuracy. Dušan Marković et al. [38] detected the presence of insects by monitoring air temperature and relative humidity. Predicted presence of insects using machine learning algorithm. 76.5% accuracy was found in the said algorithm. The detection accuracy over a five-day time interval was 86.3% and the false detection rate was 11%. Nathan Moses- The study article proposed by Gonzales et al. [39] focuses mostly on drone-based insect and pest identification. Liu Yang et al. [40] build a model to detect and classify the presence of tomato diseases and insects. Transfer Learning Algorithm Concept Using Convolutional Neural MobileNet to Classify Nine Insect Diseases in China. The model was found to be 97.5% accurate.The methodology for insect detection with deep learning follows a well-defined workflow. Tiny object detection is also a major problem in Computer vision [41]. First, a diverse collection of insect images is acquired. These images can be captured in the field or sourced from publicly available datasets. Next, the images undergo preprocessing to ensure consistency and optimal learning for the model. This might involve techniques like noise reduction, resizing, or color balancing. After that, the data is logically divided into training and testing sets. Convolutional neural networks (CNNs) use the training set as their basis for learning. The CNN carefully examines the photos as it is being trained in order to extract the traits that characterize insects. Finally, hidden data from the testing set is used to assess the model's ability. Metrics like accuracy and mean Average Precision (mAP) assess how well the trained model detects insects in new images. This iterative process of training, evaluation, and refinement allow us to continuously improve the model's accuracy for insect detection in various applications. When it comes to insect detection, our suggested model has outperformed other algorithms with remarkable results. Our model has greatly outperformed its predecessors, achieving unmatched accuracy and robustness through careful design and optimization. Using cutting-edge deep learning methods along with creative architectural improvements, in a wide range of conditions, our model identifies and classifies a wide variety of insect species with remarkable accuracy. Our suggested model's exceptional effectiveness in addressing the unique obstacles presented by insect identification jobs is one of its main advantages. Our model efficiently tackles problems like image noise, variation in insect appearance, and environmental conditions by incorporating sophisticated preprocessing techniques, guaranteeing optimal input quality for the learning process. Furthermore, our model outperforms others not just in terms of generalization and scalability but also in terms of accuracy. For The validation of the method dataset should be collected from various method and source, which should be ensured [42]

**Methods**

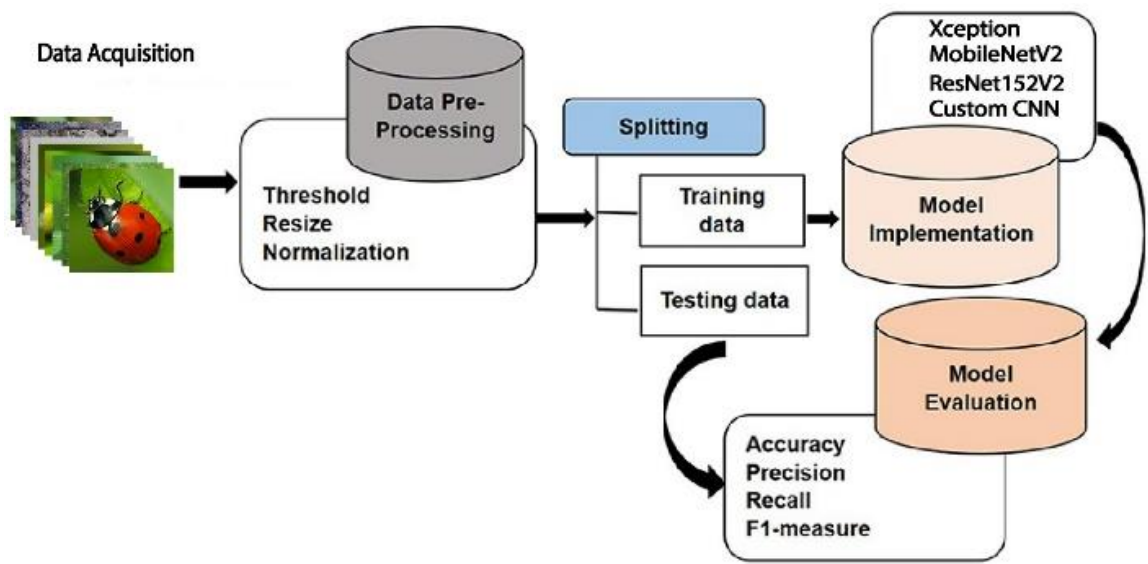

Figure 1: Architure of methodology on working process on adaptive learnig to transfer learning

## Data acquisition

The data collection phase involves carefully observing and recording specific variables related to a chosen subject, aiming to uncover patterns and correlations in the data for accurate predictions. It is typically the crucial first step in any research project, requiring a substantial amount of data, especially in machine learning where a significant dataset is necessary. In the case of identifying insect classes, images with contextual details serve as the primary data input. To quickly gather many relevant images, a Python-scripted Bot is used to automate mass image downloading. The Bot utilizes user inputs for the desired number of images and the designated file pathway. By executing background operations, the Bot retrieves the targeted images from online searches. The total dataset is almost 4509 images. The resulting dataset consists of 1200 insect images personally captured by the user and the rest obtained from Google searches using insect class keywords. Additional visuals were manually searched and obtained using the Windows screenshot tool due to the scarcity of desired images. Here show the nine classes of insect data in figure-2.

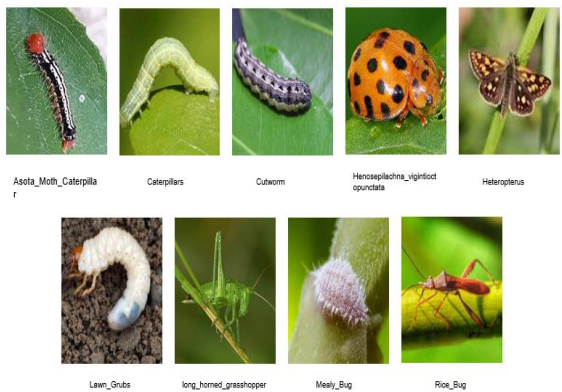

Figure-2 Dataset of insect image each class representing their different types

## Feature Extraction on Insect Image

Obtained online data, including social media, is typically unstructured. Post-data collection processing is essential to transform raw data into a usable format. This involves purging impure data, eliminating low-resolution and duplicate images, and overall cleaning the dataset. Insights and information are extracted during processing, such as generating image lists, sorting them, determining image counts per category, identifying unique images, and partitioning the dataset. Data pre-processing methods ensure dataset readiness for analysis or modeling. Steps in this project include:

a) **Resolution Conversion**- Standardizing image resolution to ensure consistency and comparability. A resolution of 448x448 was adopted for uniformity and fair comparisons.

b) **Normalization**- Scaling attribute values to address disparities and enhance model understanding. In this case, normalization between 0 and 1 was used.

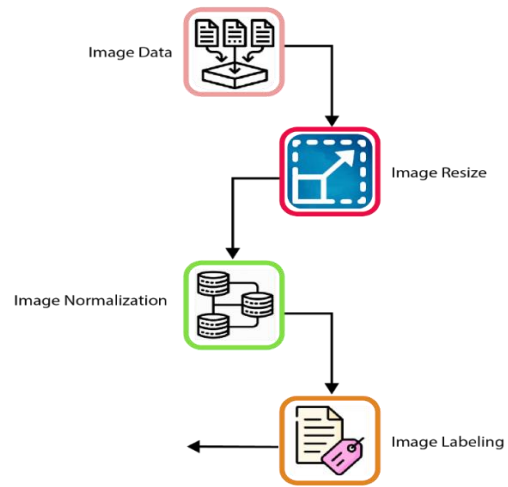

Figure-3: Preprocessing over insect images using necessary image retrieval steps

c) **Noise Removal and Image Crop:** Eliminating unnecessary noise from the picture and focusing on the ROI, which is often the skin surrounding the nail plate.

d) **Pre-processing**- datasets offers advantages such as improved results and expedited model training. Resolution conversion and normalization significantly contribute to model accuracy and training efficiency, enhancing overall performance. After preprocessing each class has 501 images.

## Distribution on Categorical Image

There are 4509 photos in the main dataset that was used in this work. The whole dataset is divided into 9 ways namely - Asota Moth Caterpillar, Caterpillars, Cutworm, Henosepilachna vigintioctopunctata, Heteropterus, Lawn Grubs, long horned grasshopper, Mealy Bug, Rice Bug. By preprocessing or cleaning the dataset, we used a fundamental partitioning of the dataset containing 100% or 4509 original documents, 80% or 3607 samples for training, 10% or 451 samples for testing and 10% or 451 samples for validation. After data preprocessing the total 4509 data are divided into 3 parts namely training, testing, validation and each part is divided into 9 parts separately. The dataset contains a total of 4509 original data, of which 501 are divided into 9 data sets.

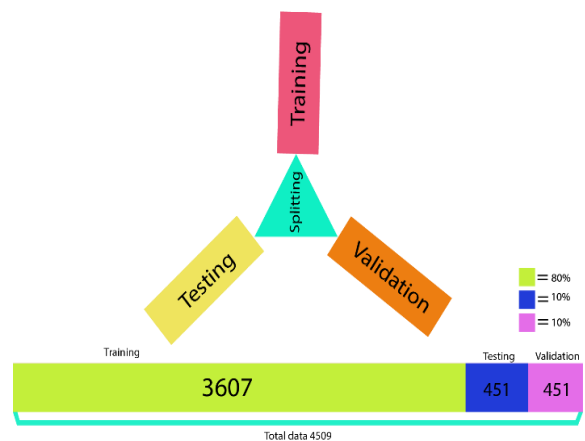

Figure-4: splitting into training, testing, validation.

## Collected Feature Learning By Adaptation technique

Data constitutes the fundamental building block of deep learning, and the pre-processing stage assumes paramount importance in the construction of any model. Following the collection and augmentation processes, all the amassed and enriched images were meticulously organized and stored locally on a drive, packaged within a compressed zip file. Access to the dataset was facilitated through the utilization of Google Colab, leveraging the directory path to establish a connection and enable seamless data retrieval and manipulation. Within the project folder, an organized structure was maintained, featuring two distinct subfolders dedicated to training and testing data. Each of these folders, in turn, encompassed an additional level of organization, with a dedicated folder created for each of the nine classes represented within the dataset. This systematic organization of the dataset ensured an efficient and streamlined workflow throughout the model development process. The clear segregation between training and testing data, along with the further categorization of classes, facilitated targeted and precise data handling, allowing for accurate training and evaluation of the deep learning model. By establishing this coherent structure and utilizing advanced techniques such as zip file storage and directory path access, the dataset management within the project environment was optimized for enhanced efficiency and seamless integration with the subsequent stages of the deep learning pipeline.

## Transfer Learning

**Model Training**: In this work, we used transfer learning to create a Convolutional Neural Network (CNN) model for the purpose of classifying insect photos. By leveraging pre-trained models like Xception, MobileNetV2, and ResNet152V2, we aimed to improve the accuracy of our custom-layered CNN model. To carefully assess each model's performance, we divided the dataset into subsets for training, testing, and validation. Through this comparative analysis, we seek to identify the most effective model for insect classification and make informed decisions for future model refinement and optimization.

**Xception**: Xception is a highly renowned computer vision model used for object detection and classification tasks. It utilizes depthwise separable convolutions, reducing computational complexity while maintaining high accuracy. Though training requires significant computational resources, its exceptional feature extraction capabilities and 94% accuracy make it widely adopted in various domains. The publicly available weights enable broad implementation and contribute to advancements in computer vision.

**MobileNetV2**: MobileNetv2 is a convolutional neural network. The MobileNetV2 model developed by Google is 53 layers deep. This model can perform real-time classification by computing on smartphone, computer, laptop devices. Currently this model outperforms MobileNetV1, SuffleNet and NASNet based on its faster estimation time, size and computational cost.

**ResNet152V2**: A Convolutional Neural Network (CNN, or ConvNet) is ResNet50V2. This deep learning model has been pre-trained to classify images. It is a branch of deep neural networks, used to implement visual image analysis. A pretrained version of the trained network can be loaded into the ResNet-50V2 ImageNet database with a total of one million images of 1000 categories.

**Custom CNN:** Convolutional Neural Network, or ConvNet, is the acronym for the CNN Algorithm. CNN is an abbreviation for Supervised Learning, a neural network subtype with a focus on data processing. There are five levels in CNN. Specifically, dropout, activation functions, fully connected layer, pooling layer, and convolutional layer.

**Proposed Model Architecture:** Strong convolutional neural network (CNN) architecture ResNet152V2 is renowned for its exceptional performance in image recognition applications. It belongs to the ResNet family of models and features 152 layers. The key innovation lies in its residual blocks, which address the vanishing gradient problem in deep networks. The inclusion of bottleneck layers reduces computational complexity, and batch normalization further enhances training efficiency. ResNet152V2's deep architecture, residual blocks, and bottleneck layers contribute to its high accuracy and computational efficiency in image recognition tasks.

**Convolutional layer:** ResNet152V2 is a deep learning model that uses a Convolutional layer with a 3x3 and 7x7 filter combo to find features and patterns in an input image. The model may capture both local and global context by using different filter sizes, which produces hierarchical representations of the image. The output consists of feature maps representing various aspects of the image, enabling accurate predictions in subsequent layers of the network.

**Residual blocks**: ResNet152V2's Residual blocks are crucial for training very deep networks and addressing vanishing gradient challenges. They introduce skip connections, enabling efficient gradient flow during training. Each block consists of stacked 3x3 convolutional layers combined with the original input via skip connections. This enables the model to concentrate on picking up the essential extra features. ResNet152V2 also uses bottleneck layers for computational efficiency. Overall, the Residual blocks enable successful training of deep networks, learning expressive representations, and mitigating gradient-related issues.

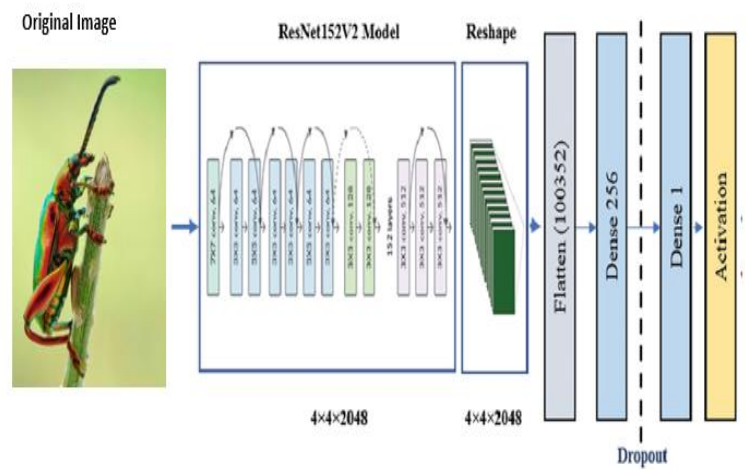

Figure-5: Model Architecture of ResNet152V2 shows it's internal connected layers with decoding layers

a) **Global average pooling**: Global Average Pooling (GAP) in ResNet152V2 aggregates spatial information from convolutional feature maps by taking the average along their height and width. This pooling operation generates a fixed-length vector that represents the most relevant and discriminative information learned throughout the network. GAP helps reduce overfitting, enhance generalization, and provides a concise representation of the input image before making final predictions.

b) **Fully connected layer:** The last part of ResNet152V2 that is in charge of converting the useful feature representation into insightful class predictions is the Fully Connected layer. Its deep layers, which are linked to each neuron in the layer above, aid in the model's comprehension of the complex relationships between traits and target classes. The Fully Connected layer uses an activation function to introduce non-linearity and is followed by a SoftMax activation for probability normalization. During training, optimization algorithms adjust the layer's weights and biases based on prediction errors. The Fully Connected layer plays a critical role in assigning class labels to input images during inference, guided by a loss function that quantifies the model's accuracy.

## Model Evaluation

Upon training a model, To evaluate its performance, two key factors need to be considered: generalization, which is represented by the test accuracy, and confidence, which is represented by the training accuracy. A measure of the model's confidence is provided by the training accuracy, which represents the model's performance on the trained dataset. But this is insufficient on its own to accurately assess the model's capabilities. The ultimate goal in model development and training is to ensure its ability to generalize well and perform accurately on new, unseen data. The term "generalization" describes the model's ability to perform well outside of the training dataset. Our datasets were split into training and testing sets, with 80% going toward training and 20% going toward testing, in order to assess generalization. The testing data was kept separate and used exclusively to evaluate the model's accuracy on unseen data throughout the training phase.

We used a standard setup with 20 steps per epoch and a learning rate of 0.011 to train each and every model. The NVIDIA GeForce RTX 3070 TI GPUs were used for the training and testing procedures. We utilized three pre-trained models, namely Xception, MobileNetV2, and ResNet152V2, to assess their performance. Among these models, ResNet152V2 exhibited the highest training and testing accuracy. In particular, ResNet152V2 scored a noteworthy 99% for training accuracy and 97% for testing accuracy. In contrast, the performance of the other models was comparatively lower than ResNet152V2. The training and testing accuracy of our transfer learning models, including ResNet152V2, are presented in Figure 1 and Table-1. These findings support ResNet152V2's superiority in terms of testing and training accuracy. Table 3 provides comprehensive information about each model's performance and shows that ResNet152V2 performs better than the other models.

Table-1: Table of Accuracy Comparison that illustrates where we can see-

| Name of Model | Accuracy of Training | Accuracy of Testing | Accuracy |
|---|---|---|---|
| MobileNetV2 | 99% | 90% | 90% |
| ResNet152V2 | 99% | 97% | 96% |
| Xecption | 97% | 92% | 94% |
| Custom CNN | 98% | 80% | 80% |

**Training and Validation Curves:**
By comparing the outcomes of the four distinct DL algorithms, it is possible to determine which one, "ResNet152V2," is operating with the maximum accuracy on the test datasets. The training and validation accuracy and loss curves for the highest accuracy model, ResNet152V2, are as follows:

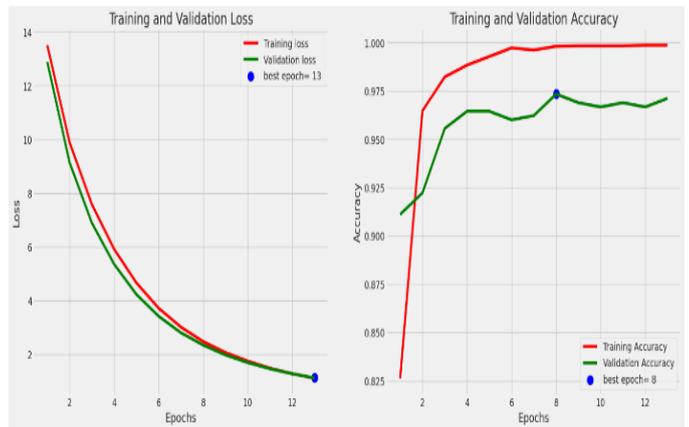

Figure-6: Classification Reports on loss and accuracy which shows the best accuracy for epoch 13

**Accuracy plot of ResNet152V2**
**Classification Report Model Performance:**

**insect image dataset:** Asota Moth Caterpillar, Caterpillars, Cutworm, Henosepilachna vigintioctopunctata, Heteropterus, Lawn Grubs, Long Horned Grasshopper, Mealy Bug, Rice Bug, by examination of the classes, categorization report, which includes f1_score, precision, recall, and support, indicated that the ResNet50v2 Model has an accuracy of 96%.

This table shows the efficacy of a deep learning model for identifying insects. Each row represents a different type of bug, like a larva of an asota moth or a cutworm. The model's performance is evaluated using three main metrics: accuracy, recall, and f1-score. Precision shows how accurate the model is in identifying a specific bug, hence reducing false positives. Recall gauges how well a model can distinguish between true and false negatives for each type of insect. Recall and precision are combined in the F1-score to provide a balanced image.

We can tell from the results that the model does a very good job overall. It attains a remarkable 96% accuracy, properly classifying 96% of the test data's bug photos.

important insights into how well the model performs in various classes. By studying these outcomes, we can identify strengths weaknesses in the model's ability to accurately classify instances within each class. While model

Table-2: Classification report of ResNet152V2 Model

| Classes | Preci-sion | Re-call | f1-score | support |
|---|---|---|---|---|
| Asota_Moth_Caterpillar | 1.00 | 1.0 | 1.00 | 48 |
| Caterpillars | 0.94 | 0.91 | 0.92 | 53 |
| Cutworm | 0.93 | 0.98 | 0.96 | 56 |
| Henosepilachna_vigintioctopunctata | 1.00 | 0.96 | 0.98 | 53 |
| Heteropterus | 0.98 | 0.96 | 0.97 | 45 |
| Lawn_Grubs | 1.00 | 1.00 | 1.00 | 64 |
| Mealy_Bug | 0.94 | 0.94 | 0.94 | 51 |
| Rice_Bug | 0.92 | 0.96 | 0.94 | 51 |
| long_horned_grasshopper | 0.91 | 0.91 | 0.91 | 46 |
| Accuracy | | | 0.96 | 449 |
| Macro avg | 0.96 | 0.96 | 0.96 | 449 |
| Weighted avg | 0.96 | 0.96 | 0.96 | 449 |

**Confusion Matrix:**

We are working with a dataset in our study that is divided into nine different classifications. We use a confusion matrix to assess our model's performance and gain insight into the categorization results. The classification results of correct classification rate and the incorrect classification rate obtained from the study of the confusion matrix offer

Table3: Confusion Matrix of ResNet152V2

| Classes | Correct Classification rate | Incorrect Classification rate |
|---|---|---|
| Asota_Moth_Caterpillar | 48 | 0 |
| Caterpillars | 48 | 5 |
| Cutworm | 55 | 1 |
| Henosepilachna_vigintioctopunctata | 54 | 2 |
| Heteropterus | 43 | 2 |
| Lawn_Grubs | 46 | 0 |
| Mealy_Bug | 48 | 3 |
| Rice_Bug | 49 | 2 |
| long_horned_grasshopper | 42 | 4 |

selection determines the minimum level of facility required to recognize insects, model evaluation checks the model's performance to see how well it can detect insects from visual data. Based on our four models, ResNet152V2 appears to provide the best outcome. We now want to evaluate this by importing random photographs to see how well our model performs. We gave this evaluation our best effort. the ResNet152V2 model. We imported the "Rice Bug" picture from the insect's picture collection.

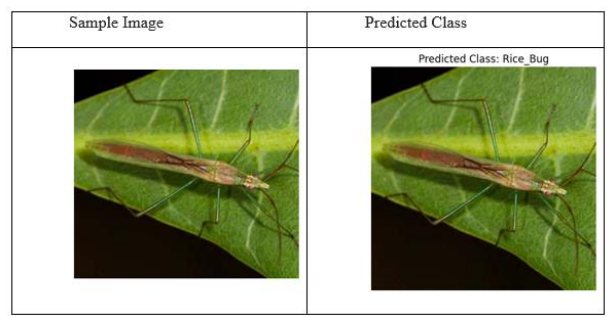

Figure-7: Rice_Bug sample image and predicted class

Therefore, we can conclude that our model will perform flawlessly if we import any real-time bug images. These findings showed that, when it comes to classifying insect photos for detection, the ResNet152V2 architecture outperforms other models (Xception, MobileNetV2).

**Conclusion**

Our findings suggest that ResNet152V2 is the best model for insect classification, with 99% training accuracy and 97% testing accuracy. Our technique provides extensive coverage of insect variety by using a painstakingly curated collection of 4,509 insect pictures, including 1,200 self-captured photographs. Automating data collecting with a Python-scripted Bot accelerates the process and improves dataset resilience. This framework offers scalable and efficient insect detection systems, which are critical for real-time monitoring and timely pest action. Our findings reflect a substantial leap in insect identification, utilizing deep learning and automation to protect agricultural output. Future advancements in AI and automation will allow stakeholders to reduce insect effect on global agriculture.

**Data Availablity**: The dataset has been published and available in public repository. doi: 10.17632/kkn8bbrgc4.1. The codes are also available in github repository: https://github.com/smshaqib/Research/tree/main/Insect%20Detection%20Models

**Declaration Statement:** The manuscript will be funded by Daffodil International University funding and Biomedical health organization. The authors declare that they have no known competing financial interests or personal relationships that could have appeared to influence the work reported in this paper.

**Authors Details:**